\documentclass[letterpaper, 10 pt, conference]{ieeeconf}  

\IEEEoverridecommandlockouts                              

\usepackage{amsmath} 
\usepackage{amssymb}  
\usepackage{graphicx}
\usepackage{booktabs}  
\usepackage{comment}  
\usepackage{mathtools}  
\usepackage{microtype}  
\usepackage{multicol}  
\usepackage{multirow}  
\usepackage{xspace} 

\usepackage{xcolor}
\definecolor{lblue}{RGB}{0, 112, 192}

\usepackage[pagebackref, breaklinks, colorlinks]{hyperref}  

\usepackage[capitalize]{cleveref}
\crefname{section}{Sec.}{Secs.}
\Crefname{section}{Section}{Sections}
\Crefname{table}{Table}{Tables}
\crefname{table}{Tab.}{Tabs.}

\newcommand{\norm}[1]{\left\lVert#1\right\rVert}

\title{\LARGE \bf
MV6D: Multi-View 6D Pose Estimation on RGB-D Frames \\Using a Deep Point-wise Voting Network
}

\author{Fabian Duffhauss$^{1}$, Tobias Demmler$^{2}$, and Gerhard Neumann$^{3}$
\thanks{$^{1}$Fabian Duffhauss is with the Bosch Center for Artificial Intelligence, Renningen, Germany, and the University of Tübingen, Germany.
        {\tt\small Fabian.Duffhauss@de.Bosch.com}}
\thanks{$^{2}$Tobias Demmler is with the Robert Bosch GmbH, Stuttgart, Germany.
        {\tt\small Tobias.Demmler@de.Bosch.com}}
\thanks{$^{3}$Gerhard Neumann is with the Karlsruhe Institute of Technology,  Karlsruhe, Germany.
        {\tt\small Gerhard.Neumann@kit.edu}}
}

\begin{document}

\maketitle
\thispagestyle{empty}
\pagestyle{empty}

\begin{abstract}

Estimating 6D poses of objects is an essential computer vision task. However, most conventional approaches rely on camera data from a single perspective and therefore suffer from occlusions. We overcome this issue with our novel multi-view 6D pose estimation method called MV6D which accurately predicts the 6D poses of all objects in a cluttered scene based on RGB-D images from multiple perspectives. We base our approach on the PVN3D network that uses a single RGB-D image to predict keypoints of the target objects. We extend this approach by using a combined point cloud from multiple views and fusing the images from each view with a DenseFusion layer. In contrast to current multi-view pose detection networks such as CosyPose, our MV6D can learn the fusion of multiple perspectives in an end-to-end manner and does not require multiple prediction stages or subsequent fine tuning of the prediction. 
Furthermore, we present three novel photorealistic datasets of cluttered scenes with heavy occlusions. All of them contain RGB-D images from multiple perspectives and the ground truth for instance semantic segmentation and 6D pose estimation. 
MV6D significantly outperforms the state-of-the-art in multi-view 6D pose estimation even in cases where the camera poses are known inaccurately. Furthermore, we show that our approach is robust towards dynamic camera setups and that its accuracy increases incrementally with an increasing number of perspectives.

\end{abstract}

\section{Introduction}

\begin{figure}[t]
  \centering 
  \includegraphics[page=1, trim = 59mm 69mm 37mm 65mm, clip,  width=1.0\linewidth]{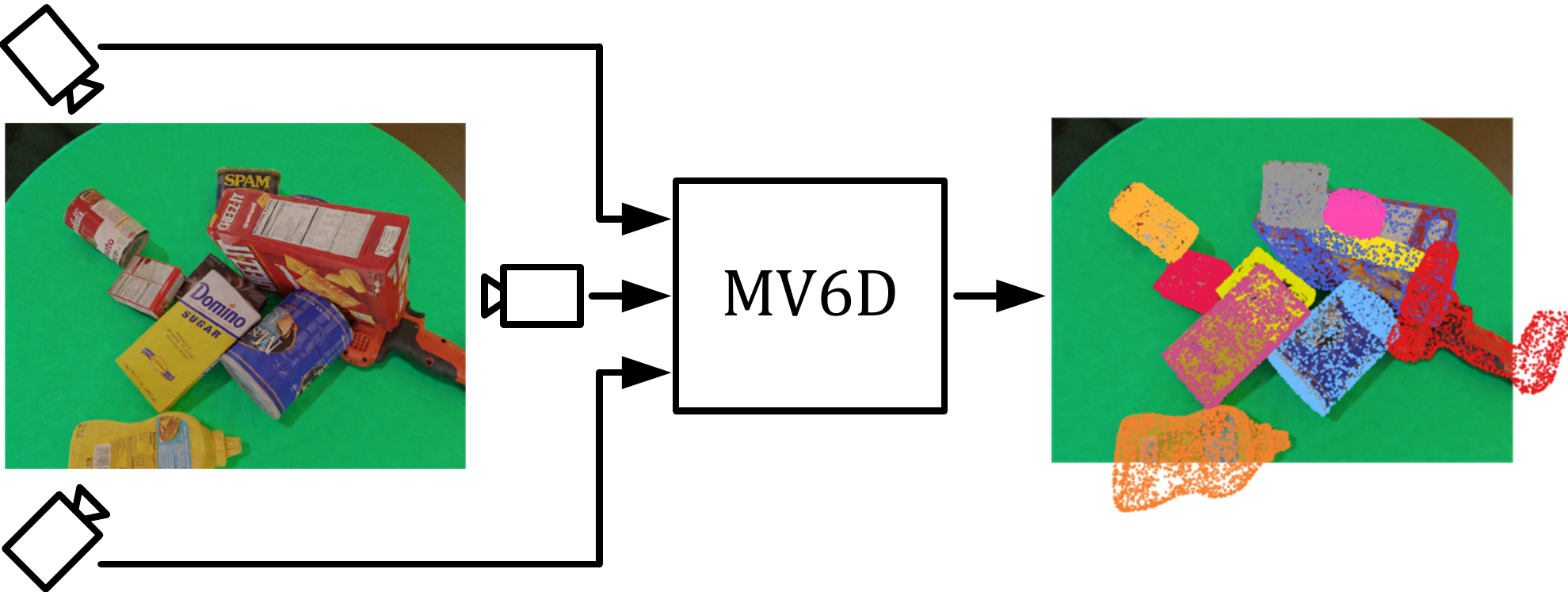}
   \caption{Overview of our MV6D approach. MV6D takes multiple RGB-D input images and predicts the 6D poses of all objects in a cluttered scene.}
   \label{fig_eye_catcher}
\end{figure}

6D pose estimation is a key technology for autonomous driving \cite{mv3d, avod, pointfusion, gu2021ecpc_icp}, 
robot manipulation \cite{moped, ffb6d, pvn3d, posecnn}, augmented reality \cite{arSurvey, su2019ar}, and human machine interaction \cite{pavlakos2017, madrigal2020robustHeadPose}. It describes the prediction of position and orientation of objects in 3D space.
Traditional pose estimation methods mostly rely on single RGB(-D) images \cite{posecnn, pvnet, densefusion, pvn3d, ffb6d} or point cloud data \cite{voxelnet, second, pointpillars}. However, these methods suffer significantly from occlusions by other objects.
To overcome this issue, we present our novel deep learning approach called MV6D which is illustrated in \cref{fig_eye_catcher}. 

MV6D takes multiple RGB-D images depicting a cluttered scene from different viewpoints which are ideally very distinct. Whereas the RGB images are processed individually, we fuse all depth images to a joint point cloud. Similar to PVN3D \cite{pvn3d} our approach predicts pre-defined 3D keypoints for each object using independent feature encoding networks for both modalities. In a final least-squares fitting \cite{leastSquares} step, we predict the 6D poses of all objects in the scene.
Since established 6D pose estimation datasets, such as YCB-Video \cite{posecnn}, LineMOD \cite{linemod}, and T-LESS \cite{tless} do not provide many frames from very distinct perspectives, we created three photorealistic datasets with cluttered scenes using YCB objects \cite{ycb}. 

We show that our multi-view approach outperforms the related single-view method PVN3D \cite{pvn3d} and the state-of-the-art multi-view 6D pose estimator CosyPose \cite{cosypose} significantly. Even in the case of inaccurate camera positioning, the accuracy of our MV6D network is much higher than the baseline methods.
Furthermore, we show that our approach copes with variable camera setups. We additionally evaluated the performance with different number of input images and show that employing a second view already leads to a large accuracy increase compared to the single-view setting while more views can further improve the results.

Our main contributions are:
\begin{itemize}
	\item We present a novel deep learning approach for 6D pose estimation on RGB-D images which learns to fuse the information from multiple views in an end-to-end manner.
	\item We present three photorealistic datasets with multi-view RGB-D data and ground truth for instance semantic segmentation and 6D pose estimation.
	\item We show that our multi-view approach outperforms the corresponding single-view approach significantly even when the positions of the cameras are known inaccurately. 
	\item Furthermore, we show that we considerably outperform current multi-view pose estimation approaches \cite{cosypose} which use a much more complex architecture including fusion of single view predictions and fine-tuning. 
\end{itemize}

\vspace{2mm}
\section{Related Work}

There has been a lot of research recently about 6D pose estimation and related tasks like 3D object detection which we present in the following.

\subsection{Pose Estimation on Single RGB Images}

Traditional pose estimation methods \cite{lowe1999sift, lowe2004distinctive, rosten2006machine, bay2006surf, rothganger2006objRec, moped, collet2009object, pvnet} extract local features from the given RGB image and match them to the corresponding features in its 3D model. Based on the 2D-3D-correspondences a Perspective-n-Point (PnP) algorithm \cite{ransac} can be applied to estimate the object's pose.
Even though feature-based methods can handle occlusions up to a certain degree, the detection of 2D keypoints does not work well on objects without a distinctive texture \cite{tless}.

In contrast, methods based on template-matching \cite{gu2010discriminative, 2011gradientResponseMaps, cao2016realTime6d} can cope with textureless objects. Templates of an object can be generated by rendering its corresponding 3D model from diverse views. Finding a match between a rendered template and a part of the input image provides the 6D pose of the corresponding object. However, template-based methods suffer from occlusions as the matching becomes inaccurate.

There are also end-to-end trainable neural networks \cite{2015viewpointsKeypoints, deepim, posecnn, gupta2019cullnet, ssd6d, tekin2018real} which directly predict the objects' poses based on a single RGB image. 

However, often the generalization of direct methods is an issue due to the non-linearity of the rotation space \cite{pvnet}. \cite{2015viewpointsKeypoints, su2015renderCNN, sundermeyer2018implicit6d} overcome this issue by discretizing the rotation space. Another common procedure is to refine the predicted poses, e.g., by applying the iterative closest point (ICP) algorithm \cite{icp} using additional depth data as in PoseCNN \cite{posecnn} or SSD-6D \cite{ssd6d}. Alternatively, deep learning-based pose refining networks like DPOD \cite{zakharov2019dpod} or DeepIM \cite{deepim} are proposed for faster and more accurate refinement without the need of depth data.

\subsection{Pose Estimation on Single Point Clouds}

Recently, due to the rapid technological progress of depth and LiDAR sensors, many pose estimation methods were developed based on a single depth image or a single point cloud \cite{chen2020survey6d, fernandes2021pointCloudSurvey}.
In this area, there are methods \cite{song2014sliding, li2017_3dFCN} that directly predict oriented 3D bounding boxes using 3D convolutions. However, 3D convolutions are computationally expensive which leads to high inference times. To reduce the computational complexity, it is common even in modern approaches \cite{voxelnet, second, pointpillars} 
to apply feature encoding networks based on PointNet \cite{pointNet} or PointNet++ \cite{pointnet++} which are able to extract geometric features. To do that, the point cloud is either divided into voxels \cite{voxelnet, second} or into vertical pillars \cite{pointpillars}. 

Recently, Qi et al. \cite{qi2019voteNet} introduced deep Hough voting for end-to-end 3D object detection. Their network VoteNet generates votes to object centers which are fused to obtain object proposals. Building upon that, Xie et al. \cite{xie2021VENet} further improved the feature encoding of seed points by an attentive multi-layer perceptron, a vote attraction loss, and vote weighting.

However, as methods based on depth images or point clouds cannot exploit texture, their performance is limited to applications where textures are not relevant.

\subsection{Pose Estimation on Single RGB-D Images}

RGB-D based approaches try to combine the advantage of both modalities. 
AVOD \cite{avod} and MV3D \cite{mv3d} use convolutional feature extractor networks followed by a 3D object proposal network for fusing RGB images and LiDAR point clouds. The latter is compressed into a bird's eye view and in case of MV3D an additional front view projection is used. However, these approaches are based on the assumption that all objects of interest are located on a plane.

PointFusion \cite{pointfusion} proposes the usage of PointNet \cite{pointNet} for point cloud feature extraction and introduces a dense fusion module for combining point cloud features and RGB features which were created by a CNN. DenseFusion \cite{densefusion} transfers that concept from 3D object detection in autonomous driving to 6D pose estimation for robotics and introduces a neural network for iterative pose refinement.

PVN3D \cite{pvn3d} further improves DenseFusion \cite{densefusion} by applying PointNet++ \cite{pointnet++} and by introducing a deep Hough voting network for 3D keypoint detection. The 6D poses are estimated by a least-squares fitting algorithm \cite{leastSquares}. FFB6D \cite{ffb6d} enhances PVN3D \cite{pvn3d} with a bidirectional fusion module that combines the features representing texture and geometric information in each encoding and decoding layer. Furthermore, they replaced PointNet++ \cite{pointnet++} by a RandLA-Net \cite{hu2020randla} for point cloud feature encoding.

\begin{figure*}[t]
  \vspace{1mm}
  \centering 
  \includegraphics[page=1, trim = 12mm 47mm 13mm 53mm, clip,  width=1.0\linewidth]{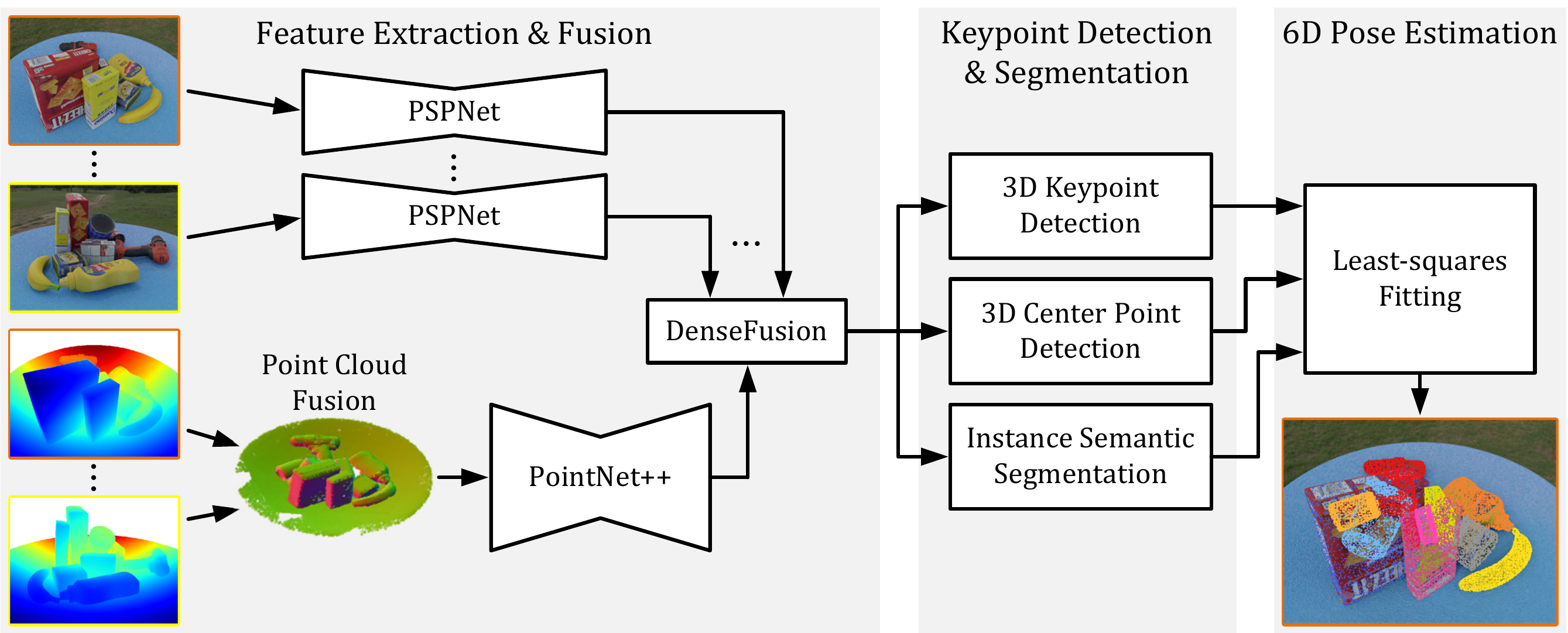}
   \caption{Architecture of our MV6D network. 
   Given multiple RGB-D images, MV6D extracts visual features from the RGB images and geometric features from a point cloud which is created by fusing all depth images. A DenseFusion network \cite{densefusion} fuses visual and geometric features. 
   Based on a 3D keypoint detection, a 3D center point detection, and an instance semantic segmentation module, we predict 6D poses using least-squares fitting \cite{leastSquares}.}
   \label{fig_architecture}
\end{figure*}

\subsection{Multi-View Pose Estimation}

Multi-view approaches aim to increase the detection accuracy by combining information from multiple perspectives.
Zeng et al. \cite{zeng2017multi} present a 6D pose estimation approach based on 15 to 18 RGB-D images that are recorded by a robot arm from very similar perspectives. They use a fully convolutional neural network to perform 2D object segmentation on each RGB image individually. Afterwards, the segmentation results are fused with the depth images into a single segmented point cloud. The 6D poses are estimated using an ICP algorithm \cite{icp}.

Sock et al. \cite{mv6dCameraMotionPlanning2017} propose an active multi-view framework with next-best-view prediction and hypothesis accumulation. Based on previous single-shot pose hypotheses they predict the next best camera perspectives and select the most likely object poses. 
Li et al. \cite{li2018unified} introduces an end-to-end trainable CNN-based architecture for 6D pose estimation based on a single RGB or RGB-D image which is used multiple times with images from different viewpoints. Afterwards, the best hypothesis is selected using a voting score that suppresses outliers. 

Recently, Labbé et al. \cite{cosypose} present an approach for 6D pose estimation based on multiple RGB images. Their method CosyPose employs DeepIM \cite{deepim} for predicting object candidates in each view separately. Subsequently, they match the object candidates that belong to the same object instance and refine the pose estimates by an object-level bundle adjustment \cite{triggs2000bundle}. However, the approach fails if an object is detected in just a single view.

All methods so far apply deep neural networks independently on each view which leads to high computational effort due to redundancy and sub-optimal use of information as there is no prediction that can use all information.
To the best of our knowledge, our approach is the first that directly fuses the features from multiple RGB-D views before performing the pose estimation based on that.

\section{Multi-View Fusion for 6D Pose Estimation}

We propose MV6D, a deep neural network for 6D pose estimation based on multi-view RGB-D data.
In the following, we define the problem of multi-view 6D pose estimation and present our deep learning approach to this.

Given an object that is observed by a camera, 6D pose estimation describes the task of predicting a rigid transformation $\boldsymbol p = [\boldsymbol R |  \boldsymbol t]$ that transforms the object from the object coordinate system into the camera coordinate system. The 6D pose $\boldsymbol p \in SE(3)$ is composed of a rotation $\boldsymbol R \in SO(3)$ and a translation $\boldsymbol t \in \mathbb{R}^3$. 
Our goal is to predict the 6D poses of all objects in a given cluttered scene based on the RGB-D images of multiple cameras with known camera pose. We assume that the 3D models of the objects are known and not more than a single instance per object class occurs in each scene.

\cref{fig_architecture} illustrates the network architecture of our approach which is composed of three stages that are inspired by the single-view network PVN3D \cite{pvn3d}.
The first stage accepts a variable number of RGB-D frames, extracts relevant features, and fuses them to a joint feature representation of the entire input scene. The second stage contains network heads for instance semantic segmentation and 3D keypoint prediction for each object. The third stage estimates the 6D poses of all objects in the scene in a least-squares fitting manner. While PVN3D uses only a single view, we propose a new mechanism to fuse the point clouds as well as the RGB images of several views into a single consistent feature representation using a modified DenseFusion module \cite{densefusion}.

\subsection{Multi-View Fusion Architecture}
\subsubsection*{Point Cloud Fusion}

To extract geometric features, we first convert all depth images into point clouds and combine them to a single point cloud using the known camera poses. As previous methods \cite{ffb6d, pvn3d}, we further process only a random subset of points 
and attach the corresponding RGB value as well as the surface normal to each remaining point. 
For feature extraction, we apply a PointNet++ \cite{pointnet++} with multi-scale grouping. 

\begin{figure*}[t]
  \vspace{1.2mm}
  \centering 
  \includegraphics[page=1, trim = 5mm 74.5mm 7mm 65mm, clip, width=1.0\linewidth]{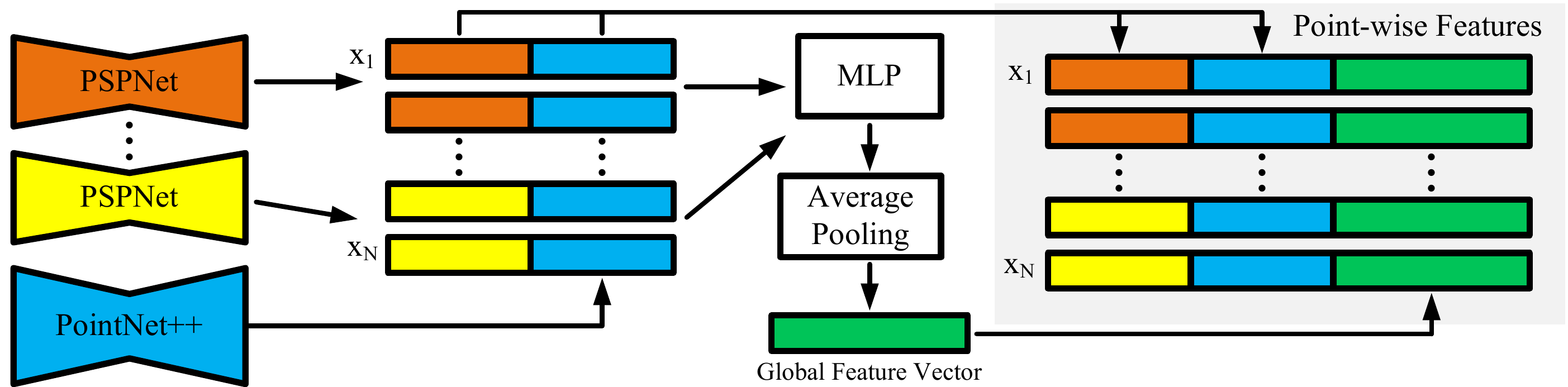}
   \caption{DenseFusion module of our network architecture. While keeping a mapping between the visual features of the PSPNets \cite{pspnet} and the corresponding geometric features of the PointNet++ \cite{pointnet++}, we compute a global feature vector and concatenate the results to obtain point-wise feature vectors.}
   \label{fig_densefusion}
   \vspace{-1mm}
\end{figure*}

\subsubsection*{RGB Image Fusion}

For each RGB image, we independently extract pixel-wise visual features using modified PSPNets \cite{pspnet} which contain a ResNet-34 \cite{resnet} pretrained on ImageNet \cite{imagenet} as backbone. All PSPNets share the same parameters. Each point in the processed point cloud is associated to a corresponding pixel of the RGB image from the view that generated the point. Similar to a DenseFusion network \cite{densefusion}, we concatenate to each point in the point cloud the PSPNet feature vector of the associated pixel from the associated view as shown in  \cref{fig_densefusion}. Hence, even if points are spatially close to each other, they can obtain the visual features from different RGB images if they have been generated from different views. 
Subsequently, we compute a global feature vector by an MLP followed by an average pooling layer that aggregates the information from the whole point cloud. This feature vector is then again appended to the geometric and RGB features of each point in the point cloud and used for further processing in the instance segmentation or keypoint detection heads.

\subsection{Network Heads}
\subsubsection*{Instance Semantic Segmentation}

The instance semantic segmentation module in \cref{fig_architecture} consists of a semantic segmentation head and a center offset head as in \cite{pvn3d}. Both submodules take the point-wise feature vectors (consisting of visual, geometric, and global features) and process them with shared multi-layer perceptrons (MLPs).
The semantic segmentation module predicts an object class for each point. The center offset module estimates the translation offset from the given point to the center of the object that it belongs to.
Following \cite{pvn3d}, we apply mean shift clustering \cite{cheng1995meanshift} to obtain the final object center predictions. These are then used to further refine the segmentation map by rejecting points which are too far away from the object center.

\subsubsection*{3D Keypoint Detection}

In advance of the training, we select eight target keypoints from the mesh of each object using the farthest point sampling (FPS) algorithm \cite{eldar1997farthest} as in \cite{pvn3d}. 
Using a shared MLP, we predict the translation offset from each point to each target keypoint of the class that the object belongs to. Adding the predicted offsets to the corresponding points from the point cloud results in a set of keypoint predictions. 
All keypoint predictions belonging to an instance are clustered by mean shift clustering \cite{cheng1995meanshift} in order to obtain the final 3D keypoint predictions as in \cite{pvn3d}.

\subsection{Loss Function}

The network is trained as in \cite{pvn3d} by minimizing the multi-task loss function 
\begin{equation}
    L_\text{multi-task} = \lambda_1 L_\text{keypoints} 
+ \lambda_2 L_\text{semantic}  
+  \lambda_3 L_\text{center},
\end{equation}
where the 3D keypoints detection loss $L_\text{keypoints}$ and the center voting loss $L_\text{center}$ are L1 losses, whereas the semantic segmentation loss $L_\text{semantic}$ is a Focal loss \cite{focalLoss}.
$\lambda_1 = \lambda_3 = 12$ and $\lambda_2 = 1$ are manually chosen weights for the individual loss functions. These weights correspond to the reference implementation of He et al. \cite{pvn3d} on GitHub but not to the values they published in their paper (i.e., $\lambda_1 = \lambda_2 = \lambda_3 = 1$).
In their code, they actually add keypoint and center point losses for the entire batch using batch size 24 while averaging the segmentation loss over the entire batch. We made the implementation independent of the batch size by averaging over all losses individually which explains the high values for $\lambda_1$ and $\lambda_3$. We found that, in this way, the optimization was easier to tune, for example, making choice of the $\lambda$ parameters and of the learning rate independent of the batch size.

\subsubsection*{Least-squares Fitting}

Based on the target keypoints and the corresponding keypoint predictions, we use least-squares fitting \cite{leastSquares} to compute the rotation $\boldsymbol R$ and the translation $\boldsymbol t$ of each object following \cite{pvn3d}.

The least-squares fitting algorithm \cite{leastSquares} determines rotation $\boldsymbol R$ and translation $\boldsymbol t$ of the 6D pose by minimizing the following squared loss as in \cite{pvn3d}
\begin{equation}
    L_\text{Least-squares} = \sum_{i=1}^M \norm{\boldsymbol{\widehat{k}_i} - (\boldsymbol R \boldsymbol k_i + \boldsymbol t)}_2^2,
\end{equation}
where $M=8$ is the number of keypoints per object, $\boldsymbol k_i$ are the target keypoints in the object coordinate system, and $\boldsymbol{\widehat{k}_i}$ are the predicted keypoints in the camera coordinate system.

\vspace{1.5mm}
\section{Experimental Setup}
\label{sec:experimentalSetup}

\subsection{Datasets}

The most common datasets for 6D pose estimation are YCB-Video \cite{posecnn}, LineMOD \cite{linemod}, Occlusion LineMOD \cite{occLinemod}, and T-LESS \cite{tless}.
Unfortunately, all of these datasets provide annotated training data for only a few different scenes, e.g. 13 scenes in LineMOD, 80 scenes in YCB-Video, and just single-object training data in T-LESS. Besides, YCB-Video and LineMOD do not show a high amount of clutter and the scenes are only shown from similar perspectives instead of clearly distinct views. Therefore, these datasets are not suitable to train and evaluate our proposed method which is designed for very distinct views in heavily cluttered scenes.

\subsubsection*{Novel Photorealistic Datasets}

Due to the drawbacks of the previously mentioned datasets, we created three novel photorealistic datasets of diverse cluttered scenes with heavy occlusions. All of these datasets contain RGB-D images from multiple very distinct perspectives which are annotated with 6D poses of all cameras and objects as well as ground truth for instance semantic segmentation.

Our datasets are composed of eleven non-symmetric objects from the YCB object set \cite{ycb}. 
Many of these objects have a symmetric shape but are non-symmetric due to their texture. This requires the pose estimation approaches to exploit RGB information in order to correctly predict the objects' poses.
Using Blender with physics, we created cluttered scenes by spawning the YCB objects above the ground in the center with a 3D normally distributed offset. Due to the similar spawning point of all objects, it is likely that objects rest on top of each other. Instead of a flat ground plane, we use a shallow bowl for the objects to fall onto. The bowl prevents the objects from scattering in all directions. Consequently, the objects stay close together resulting in heavily occluded scenes.

In addition to the random composition of the scenes, we applied further domain randomization techniques in order to improve the generalization of our approach. For each scene, we selected a random $360\times180$ degree panorama photo from a set of 379 photos and a different bowl texture. Furthermore, we slightly vary intensity and color of the lighting to create different shadows and reflections.

For our three datasets, we collected RGB-D images from multiple views with different camera settings. Therefore, we call our dataset framework Multi-View YCB (MV-YCB). We generated ground truth for instance semantic segmentation and 6D pose estimation. Additionally, we provide the exact camera poses. All datasets are split into 90\% training data and 10\% test data.

For the first dataset, called MV-YCB FixCam, we generated 8,333 random scenes and placed three identical cameras at fixed positions equally distributed around the scene, i.e. in a circle in 120 degree intervals. This results in a total of 24,999 RGB-D frames with corresponding ground truth data.

For the second dataset, called MV-YCB WiggleCam, we use the same scenes and cameras as in MV-YCB FixCam, but add a 3D normally distributed offset to each camera independently. This reflects a typical robotic setting where the cameras are mounted slightly inaccurate around the scene.

For the third dataset, called MV-YCB MovingCam, we generated further 8,333 random scenes and placed four cameras around the scene, where each camera spawns in another quadrant in a sphere. This represents a scenario, where a single camera is moved around the scene recording four frames from four very distinct perspectives. This results in a total of 33,332 RGB-D frames with corresponding ground truth data.

\subsection{Data Augmentation}
\label{sec_augmentations}

We used a variety of data augmentations to boost the generalisation of our network. During training we followed \cite{pvn3d} and applied a random combination of color jitter, sharpening filters, motion blur, Gaussian blur, and Gaussian noise on the RGB images. The depth maps and the corresponding point clouds are not augmented.

\subsection{Training Procedure}

For MV-YCB FixCam and MV-YCB WiggleCam, we train with all three camera views and increase the number of training samples by using all relevant camera combinations i.e. \{[cam1, cam2, cam3], [cam2, cam3, cam1], [cam3, cam1, cam2]\}. We do not need every possible permutation of this list since the exact order is irrelevant. Only the first camera in the list determines the camera coordinate system in which the 6D poses are predicted.

For MV-YCB MovingCam, we use a variable number of camera views.
It allows for more flexibility when deploying the trained network as the network learns how to deal with a different number of views.
We employ a set with all relevant camera combinations for each view count, e.g. \{[cam1, cam2], [cam1, cam3], [cam2, cam3]\} for a view count of two if we use a maximum of three cameras. The previous rotation step is then applied on each item. 

Each sample in a batch must have the same view count. When training with a variable view count this is ensured by using a custom batch sampler. An equal amount of batches is sampled for each view count. A sample with a lower view count requires less memory and offers less information for network optimization. Therefore, we sample more samples per batch for lower view counts in order to further balance the optimization. Consequently, each batch has roughly the same memory size and information amount.

\subsection{Implementation Details}

For the training on MV-YCB FixCam and MV-YCB WiggleCam, we used eight GPUs of type NVIDIA Tesla V100 with 32GB of memory. We randomly sample 12,288 points from each depth map. 
For the training on MV-YCB MovingCam, we used three NVIDIA GeForce RTX 2080 Ti with 11GB of memory. Here we sample 6,144 points from each depth map. 
All networks were trained with an Adam optimizer \cite{adam} using a cyclical learning rate \cite{cyclicLR} with the triangular2 policy.

\subsection{Evaluation Metrics}
\label{sec_eval_metrics}

For evaluating our network and comparing it with other approaches, we follow previous works \cite{pvn3d, ffb6d, densefusion, posecnn} and use the area-under-curve (AUC) metrics for ADD-S and ADD(-S) as well as the percentage of predictions ADD-S and ADD(-S) \textless ~2cm.

The average distance metric ADD \cite{hinterstoisser2012model} is computed by
\begin{equation}
    \text{ADD} = \frac{1}{|\mathcal M|} \sum_{\boldsymbol x \in \mathcal M} \norm{(\widehat{\boldsymbol R} \boldsymbol x + \widehat{\boldsymbol t}) - (\boldsymbol R \boldsymbol x + \boldsymbol t)}_2 ,
\end{equation}
where $\mathcal M$ is the set of vertices of a given object mesh. $\boldsymbol p = [\boldsymbol R | \boldsymbol t]$ and $\widehat{\boldsymbol p} = [\widehat{\boldsymbol R} | \widehat{\boldsymbol t}]$ are the ground truth pose and the predicted pose respectively. The average distance is calculated between all corresponding vertices of the ground truth pose and the predicted pose.

The average closest point distance metric ADD-S \cite{hinterstoisser2012model} is a slightly relaxed adaptation of the ADD metric which is more suitable for symmetric objects. It is computed by
\begin{equation}
    \text{ADD-S} = \frac{1}{|\mathcal M|} \sum_{\boldsymbol x_1 \in \mathcal M} \min_{\boldsymbol x_2 \in \mathcal M} \norm{(\widehat{\boldsymbol R} \boldsymbol x_1 + \widehat{\boldsymbol t}) - (\boldsymbol R \boldsymbol x_2 + \boldsymbol t)}_2.
\end{equation}
Here, the average distance is computed between each vertex of the predicted pose and the closest vertex on the ground truth.

The average (closest point) metric ADD(-S) \cite{pvn3d} is a combination of the previous two metrics. If it is applied on a symmetric object, the relaxed ADD-S metric is used. If the object is non-symmetric, the stricter ADD metric is applied. 

Based on the values of ADD-S and ADD(-S), we compute the area under the accuracy-threshold curve (AUC) and the percentage that is smaller than 2cm which is a typical threshold for successful robot manipulation.

\subsection{Baseline Methods}
\label{sec_baseline_methods}

We compare our approach with the single-view performance of PVN3D \cite{pvn3d} (which was the best 6D pose estimator when we started this project) and the best multi-view 6D pose estimator CosyPose \cite{cosypose}. Originally, CosyPose is based on the single-view approach DeepIM \cite{deepim}, uses only RGB data, and can cope with unknown camera poses. 
However, CosyPose is already outperformed by PVN3D on LineMOD \cite{linemod} and YCB-Video \cite{posecnn} as both papers suggest \cite{cosypose, linemod}. In order to get a more challenging benchmark than PVN3D, we exchanged the DeepIM in CosyPose with PVN3D in order to create a multi-view 6D pose estimator based on RGB-D data. Since our approach assumes the camera poses to be known, we evaluated CosyPose not only with unknown camera poses as proposed in the paper \cite{cosypose} but also used the known camera poses in a second evaluation instead of their estimations to improve the hypothesis matching and to make the comparison with our approach fairer.

\vspace{1mm}
\section{Results}
\label{sec:results}

\subsection{Quantitative Results on MV-YCB FixCam}

\cref{tab_fixCam} shows the 6D pose estimation accuracy of our MV6D network using the metrics presented in \cref{sec_eval_metrics}. It is compared with the single-view performance of PVN3D \cite{pvn3d} and the multi-view performance of CosyPose \cite{cosypose} as described in \cref{sec_baseline_methods}. The best results are printed in bold.

We can see that MV6D outperforms PVN3D and CosyPose on all metrics significantly. Using the known camera poses instead of estimating them leads to a small improvement on CosyPose but its accuracy stays significantly lower that ours. Even though the MV-YCB FixCam dataset has heavy occlusions, more than 96\% of all objects can be predicted within the 2cm robot manipulation threshold. 

\begin{table}[tb]
    \centering
    \vspace{1.5mm}
    \caption{Quantitative results on MV-YCB FixCam}
    \label{tab_fixCam}
\begin{tabular}{r|cccc}
    \toprule
                       & PVN3D      & CosyPose  & CosyPose   &  MV6D           \\
Number of views        & 1          &  3        &  3         &  3              \\
Known camera poses     & \checkmark & $\times$  & \checkmark &  \checkmark     \\
\midrule
ADD-S AUC              &   81.3     &   90.8    &    91.9    & \textbf{96.9}  \\
ADD(-S) AUC            &   74.9     &   82.4    &    84.6    & \textbf{94.8}  \\
ADD-S \textless   ~2cm &   82.1     &   92.9    &    93.0    & \textbf{98.8}  \\
ADD(-S) \textless ~2cm &   73.0     &   80.6    &    82.4    & \textbf{96.5}  \\
\bottomrule
\end{tabular}
\vspace{-0.5mm}
\end{table}

\subsection{Quantitative Results on MV-YCB WiggleCam}

\cref{tab_wiggleCam} presents the results on the MV-YCB WiggleCam dataset where the known camera poses deviate slightly from the actual camera poses.

We see that the inaccurate camera positioning leads to a small accuracy decrease of all approaches.
Nevertheless, our multi-view network still outperforms all other methods significantly.

\begin{table}[tbh]
    \vspace{1.5mm}
    \centering
    \caption{Quantitative results on MV-YCB WiggleCam}
    \label{tab_wiggleCam}
\begin{tabular}{r|cccc}
    \toprule 
                        & PVN3D      &  CosyPose &  CosyPose  &  MV6D         \\
Number of views         & 1          &  3        &  3         &  3            \\
Known camera poses      & \checkmark & $\times$  & \checkmark &  \checkmark   \\
\midrule
ADD-S AUC               & 80.8       &  90.0     &   91.3     & \textbf{96.2} \\
ADD(-S) AUC             & 74.0       &  81.0     &   83.4     & \textbf{93.0} \\
ADD-S \textless   ~2cm  & 82.0       &  92.3     &   92.6     & \textbf{98.7} \\
ADD(-S) \textless ~2cm  & 72.4       &  78.9     &   81.6     & \textbf{96.0} \\
\bottomrule
\end{tabular}
\vspace{-0.5mm}
\end{table}

\subsection{Quantitative Results on MV-YCB MovingCam}

\cref{tab_auc_movingCam} shows the AUC results of PVN3D \cite{pvn3d} and MV6D on the MV-YCB MovingCam dataset for each object class individually.
In this scenario, we trained MV6D with a varying number of views and evaluated the same model on one to four views. 

The high accuracy of MV6D proves that our architecture copes very well with the dynamic camera set\-up. MV6D outperforms PVN3D vastly even when using just two views and the accuracy further increases with an increasing number of input images. Furthermore, MV6D almost matches PVN3D when inputting just a single view which is naturally not the designated use case of our method.

\begin{table*}[h]
    \centering
    \vspace{2mm}
    \caption{AUC results on MV-YCB MovingCam. The best results are printed in bold.}
    \label{tab_auc_movingCam}
\begin{tabular}{l|cc|cc|cc|cc|cc}
    \toprule
\multirow{3}{*}{Object classes} & \multicolumn{2}{c|}{PVN3D} & \multicolumn{8}{c}{MV6D (variable number of views)} \\
  & \multicolumn{2}{c|}{1 view} & \multicolumn{2}{c|}{1 view} & \multicolumn{2}{c|}{2 views} & \multicolumn{2}{c|}{3 views} & \multicolumn{2}{c}{4 views} \\
				   & ADD-S &ADD(-S)& ADD-S &ADD(-S)& ADD-S &ADD(-S)& ADD-S 			& ADD(-S) 		 & ADD-S  		  & ADD(-S)        \\
\midrule
Banana             & 80.9 & 73.0 & 80.5 & 73.0 & 94.3 & 90.1 &         96.7  &         94.2  & \textbf{97.5} & \textbf{95.9} \\
Cracker box        & 97.2 & 96.8 & 96.9 & 96.3 & 97.9 & 97.7 & \textbf{98.2} &         98.0  & \textbf{98.2} & \textbf{98.1} \\
Gelatin box        & 78.1 & 73.4 & 76.3 & 72.2 & 93.4 & 90.4 &         96.5  &         94.6  & \textbf{97.7} & \textbf{96.4} \\
Master chef can    & 94.5 & 91.5 & 94.6 & 91.4 & 98.1 & 97.1 & \textbf{98.2} & \textbf{97.7} & \textbf{98.2} & \textbf{97.7} \\
Mustard bottle     & 91.2 & 87.2 & 90.2 & 86.8 & 97.2 & 95.5 &         97.7  &         96.9  & \textbf{97.9} & \textbf{97.2} \\
Potted meat can    & 88.0 & 84.9 & 86.9 & 83.7 & 97.2 & 96.1 &         98.3  &         97.6  & \textbf{98.4} & \textbf{97.9} \\
Power drill        & 94.4 & 92.6 & 93.6 & 91.5 & 97.1 & 96.3 &         97.2  &         96.8  & \textbf{97.8} & \textbf{97.4} \\
Pudding box        & 86.4 & 82.6 & 85.4 & 82.0 & 95.8 & 94.3 &         97.7  &         96.8  & \textbf{98.3} & \textbf{97.6} \\
Sugar box          & 93.1 & 90.5 & 91.9 & 89.5 & 97.5 & 96.7 &         98.0  &         97.6  & \textbf{98.1} & \textbf{97.8} \\
Tomato soup can    & 87.7 & 83.6 & 87.3 & 83.3 & 97.3 & 95.4 &         98.2  &         97.2  & \textbf{98.4} & \textbf{97.7} \\
Tuna fish can      & 78.6 & 71.6 & 77.3 & 70.8 & 93.1 & 88.5 &         97.5  &         94.5  & \textbf{97.6} & \textbf{95.3} \\
\midrule      
ALL                & 88.2 & 84.4 & 87.4 & 83.7 & 96.3 & 94.4 &         97.7  &         96.5  & \textbf{98.0} & \textbf{97.2} \\
\bottomrule
\end{tabular}
\end{table*}

\subsection{Qualitative Results}

\begin{figure*}[ht]
    \centering
    \begin{minipage}{.248\textwidth}
    \centering
    \textbf{\large Ground-Truth}
    \end{minipage}%
    \begin{minipage}{.248\textwidth}
    \centering
    \textbf{\large PVN3D}
    \end{minipage}%
    \begin{minipage}{.248\textwidth}
    \centering
    \textbf{\large CosyPose}
    \end{minipage}%
    \begin{minipage}{.248\textwidth}
    \centering
    \textbf{\large MV6D}
    \end{minipage}%
    \vspace{1.5mm}
    \setkeys{Gin}{width=0.248\linewidth}
    \includegraphics{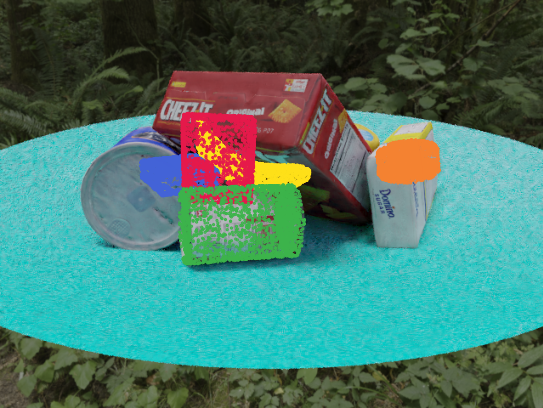}\,%
    \includegraphics{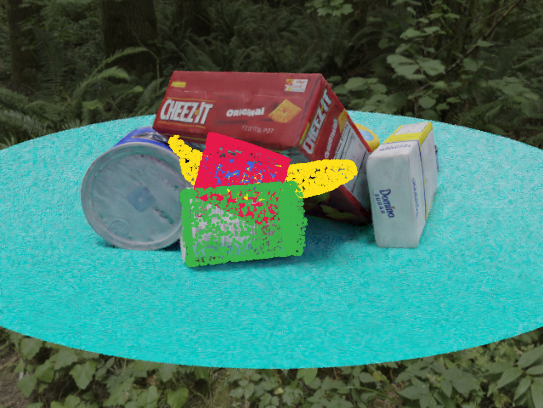}\,%
    \includegraphics{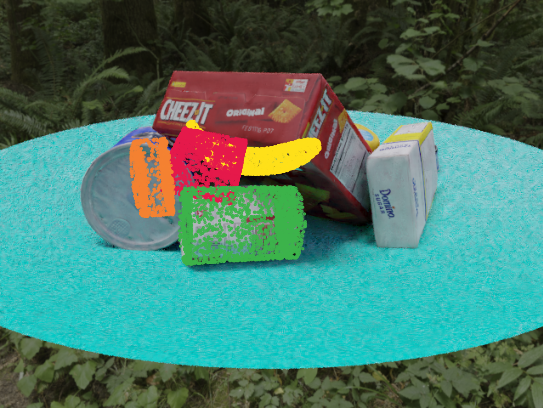}\,%
    \includegraphics{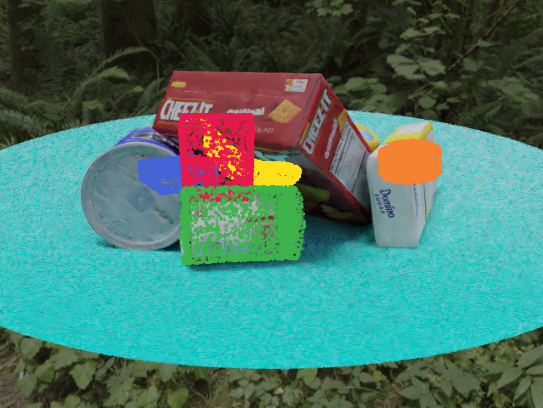}
    \vspace{-2.7mm}
    
    \includegraphics{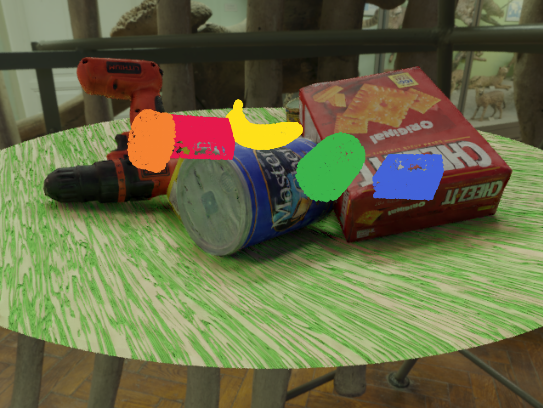}\,%
    \includegraphics{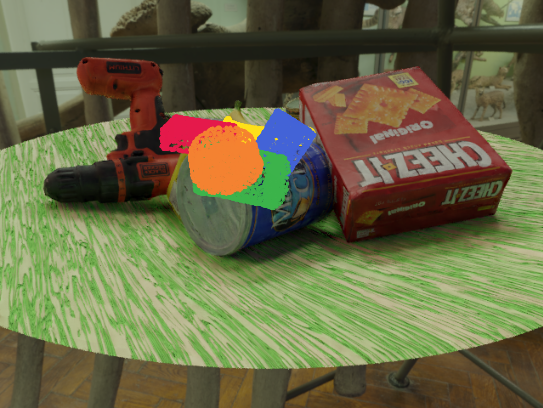}\,%
    \includegraphics{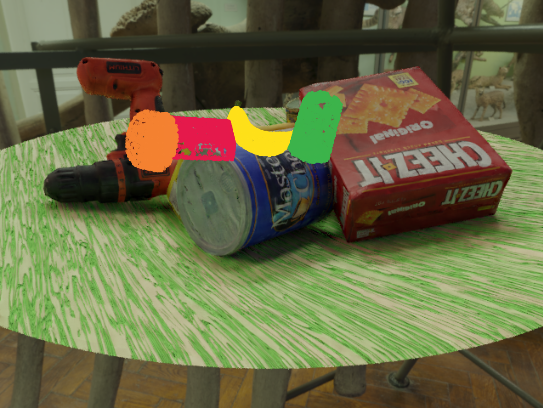}\,%
    \includegraphics{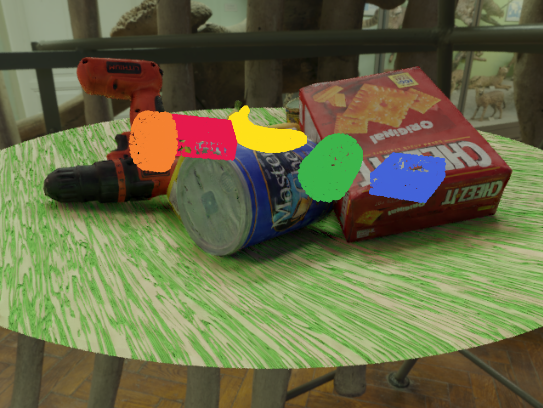}
    \vspace{-2.7mm}
    
    \includegraphics{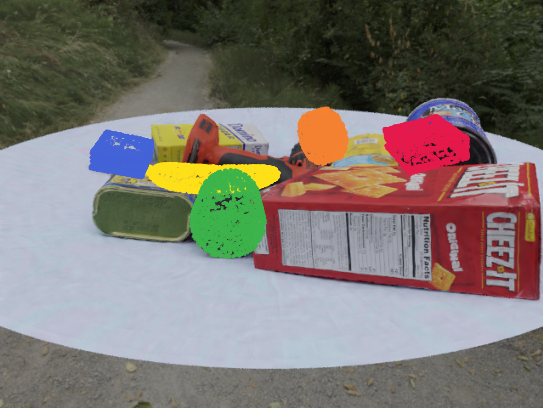}\,%
    \includegraphics{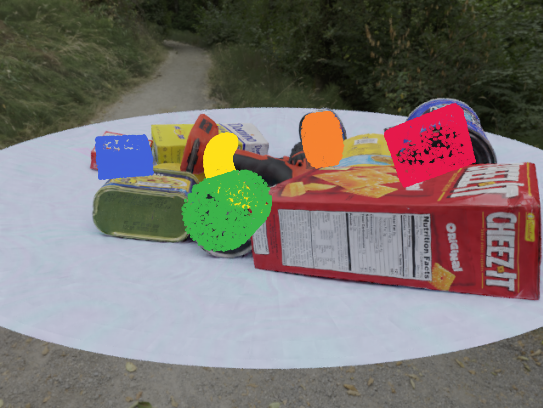}\,%
    \includegraphics{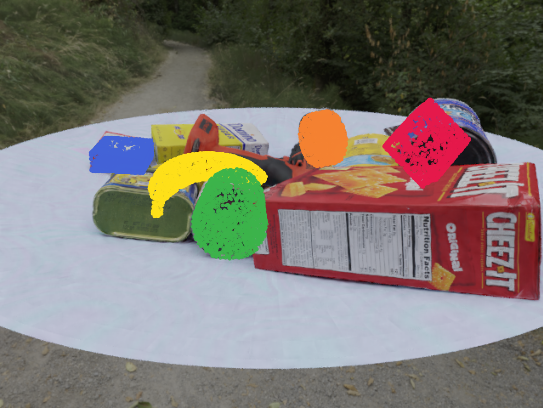}\,%
    \includegraphics{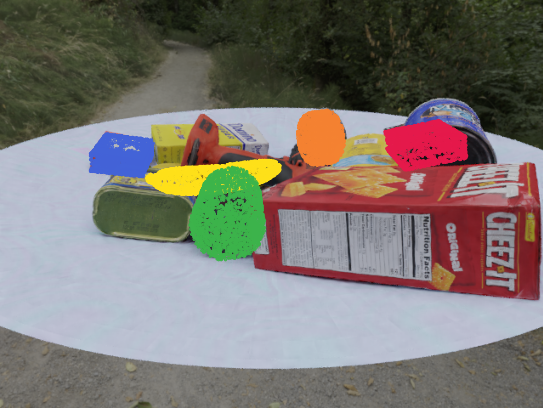}
\caption{6D pose predictions of PVN3D \cite{pvn3d}, CosyPose \cite{cosypose}, and MV6D in comparison to the ground-truth on MV-YCB FixCam. The three rows show three different example scenes that represent the typical performance of the networks. For clarity, only the poses of the five most difficult objects are visualized: tuna fish can (orange), banana (yellow), tomato soup can (green), gelatin box (blue), and pudding box (red).}
\label{fig_fixCam_grid}
\end{figure*}

\cref{fig_fixCam_grid} shows qualitative results of our MV6D network on the MV-YCB FixCam dataset in comparison to PVN3D \cite{pvn3d}, CosyPose \cite{cosypose}, and the ground-truth.
We can see that our approach predicts the 6D poses of all objects very accurately even though some objects are heavily occluded. 
As PVN3D gets just the single RGB-D image from the depicted view, it cannot detect some objects at all, such as the tuna fish can and the gelatin box in the first row. CosyPose, which uses the PVN3D predictions on all three views, performs therefore usually better than PVN3D, but for heavily occluded objects, MV6D still outperforms it.

\section{Conclusion}

We have presented MV6D, a deep learning approach for multi-view 6D pose estimation based on RGB-D data. Given multiple RGB-D recordings of a cluttered scene consisting of multiple known objects, our approach fuses the depth maps into a joint point cloud. The RGB images are processed independently by a CNN before their features are fused with the point cloud features. Based on the fused features, we predict 3D keypoints, 3D center points, and semantic labels, which are used to estimate the objects' 6D poses using a least-squares fitting algorithm.

For examining the strengths of our MV6D network, we created three novel photorealistic datasets with cluttered scenes recorded by multiple RGB-D cameras from different perspectives. Our experiments prove that our multi-view approach outperforms 
the baseline methods 
by a large margin even when the camera positions are known only inaccurately.
Moreover, we prove that our method masters dynamic camera settings and that its accuracy incrementally rises with an increasing number of input images.




{\small
\bibliographystyle{IEEEtran}
\bibliography{IEEEabrv}
}

\end{document}